\documentclass[letterpaper, 10 pt, conference]{ieeetran}  

\IEEEoverridecommandlockouts                              

\usepackage{graphicx}
\usepackage{amsmath} 
\usepackage[normalem]{ulem}
\usepackage[flushleft]{threeparttable}
\usepackage{eso-pic}

\title{\LARGE \bf
On the Formal Development of Behavioral Reactive Agents: A Systematic Braitenberg-Vehicle Approach
}

\author{Matin Macktoobian$^{1}$ and Ahmad KhatamiNejad T.$^{2}$
\thanks{$^{1}$Matin Macktoobian is with Systems Control Group, The Edward S. Rogers Sr. Department of Electrical \& Computer Engineering, University of Toronto, Ontario, Canada,
        {\tt\small matin.macktoobian@scg.utoronto.ca}}%
\thanks{$^{2}$Ahmad KhatamiNejad T. is with the Image Processing Lab, Department of Computer Engineering, Sharif University of Technology, Tehran, Iran,
        {\tt\small khataminejad@ce.sharif.ac.ir}}%
}


\begin{document}
\IEEEoverridecommandlockouts                             



\maketitle

\begin{abstract}

In this paper, a novel process has been developed to realize high-level complex cognitive behaviors into reactive agents, efficiently. This method paves the way for deducting high-level reactive behaviors from low-level perceptive information by autonomous robots. The aforementioned process lets us actualize different generations of Braitenberg vehicles, are which able to mimic desired behaviors to survive in complex environments with high degrees of flexibility in perception and emergence of high-level cognitive actions. The approach has been used to engineer a Braitenberg vehicle with a wide range of perception-action capabilities. Verification would be realized within this framework, due to the efficient traceability between each sequential pair of process phases. The applied simulations demonstrate the efficiency of the established development process, based on the Braitenberg vehicle's behavior.

\end{abstract}


\section{INTRODUCTION}

Braitenberg vehicles(BV) \cite{1} have played a central role in identification of cognitive systems and led to immense progression toward controllability of reactive agents in robotics. Utilization of neural approaches, specially spiking neural networks(SNN), to model dynamics of BVs have arranged an efficient framework to model and analyze the functional behavior of the agents. A coherent survey for reviewing the neural dynamics of SNNs and spike transitions in SNNs could be found in \cite{2}.

Animating low-level neural vehicles to perceive and behave within the scope of the local environment, reactively, has shaped this track of research in cognitive robotics. Regulation of the neural transitions and a learning-based weight manipulating approach have been proposed in \cite{3}. It has utilized neuromodulation concept within a neural network to control a typical mobile agent with some specific reactive characteristics. However, exclusively-modeling of foraging behavior with neuromodulation and lack of any implication about the generality of this method do cast doubt on the applicability of the approach to control the other types of behaviors and behavioral mobile agents.

\cite{4} conveys a primary but pioneering endeavor to emulate the cognition and intelligence by agents. Utilization of the heuristics might be stressed to consider some degrees of proactivity in that context. Described design principles for autonomous agents by that research are, heavily, dependent on the heuristics. One could contend that the predicting feature of heuristics do not fit well in control of reactive agents, like BVs; Because, any perception-action cycle shall be done with neither any explicit memorized event nor implications about the future. Furthermore, this research overview regarding the design is, noticeably, high-level and far apart a process-oriented framework to develop autonomous agents, albeit still efficient requirements to underpin any potential development framework. Our approach would be depicted as an evolved implementable derivation of the requirements, have which been proposed in this paper. Note that the development by aforesaid requirements in the context of reactive agents could lead to their immense customization. The outcomes and qualitative aspects of such advancements shall be investigated in this research.

Promotion of BVs to perceive more complicated proximity patterns has just been coined with \cite{5}, within which bidirectional straight motion have been, successfully, perceived. Mobilized reactive agents with curved trajectory detection (CTD) mechanism, in both directions, have been addressed in \cite{6}. The proposed SNN-based circuit does, actually, realize the detection of major 2D planar trajectories around a typical agent. The neural-cross-correlative mechanism governing CTD \cite{7} does, implicitly, exhibit that presentation of complicated perceiving circuits should be taken into account by a process to both decrease the design complexities and pave the way of systematic verification.

One might trace more various proposed control schemes in this context. As an instance, \cite{8} has presented a study on the stochastic behavior of a typical BV and potential advancements to control the agent. Moreover, utilization of spatial representation for reasoning and control could be named as the other useful methodologies to handle BVs and cognitive robots, efficiently  \cite{9} \& \cite{10}.

Prior work in study of spatiotemporal behaviors, is which dedicated to the BVs, has focussed on extraction of spatiotemporal knowledge from the same type of perception \cite{11}. This scheme, fundamentally, regards the local communication assumption among the nearby agents. Evolution of the agents' virtual representation from the outer world does drive us to specify the method as an semi-supervised learning protocol.

Most recent approaches have, relatively, focussed on architectural and adaptive ideas to control and improve reactive agents. A control architecture based on the engineering paradigms corresponding to software systems for accelerating the decision making task for reactive systems has been investigated in \cite{12}. A symbolic variant of dynamic programming approaches \cite{13} has been used to reactive planning of agents. The promising denouements of this research reinforce the usability of this method in open-world scenarios of agent planning. Furthermore, one could mention invention of the bilinear method \cite{14} to process high-level perceptive environmental information for acquisition of learning and adaptivity in life time of a typical agent. Behavior composition problem \cite{15}, also, focuses on mimicking a target behavior by sequential execution of some detecting modules, where perceiving modules would convey high-level information about the environment \cite{16}. A similar point of view could be studied in \cite{17}, within which infinite sets of acceptable behaviors will be generated for the agents, albeit acceded to acquire partially-controllable behaviors. The endeavor to optimize the solutions for behavior composition problem \cite{18} has been led to inspiring insights in the definition of fully-realizable behaviors.

This paper is going to establish a process-oriented mechanism to design reactive agents and BVs, according to desired behavior for them. The approach aims to handle complexity of the target behavior in the phase of design and implementation, as the processing phases shall be taken into account in either simple or complicated design patterns. This formal developing strategy does provide the transparent view of the traceability between required behavior and the actuation capabilities of the agent. Obviously, verification could be applied, effectively, onto this context to devise a robust structure to assess the behavioral characteristics of the candidate BV. The paper has been organized, to cover the target contribution, as follows: Overall specification of the process shall be investigated on section \ref{section_2}. Section \ref{section_3} presents Opportunist, as a novel highly-cognitive BV, is which developed by our novel process. The analysis of the Opportunist functionalities and the accomplishments have been reviewed in section \ref{section_4}. Section \ref{section_5} depicts the conclusion and potential directions of the research, based on the proposed process and its applications.

\begin{figure}[!t]
\centering
\includegraphics[scale=0.5]{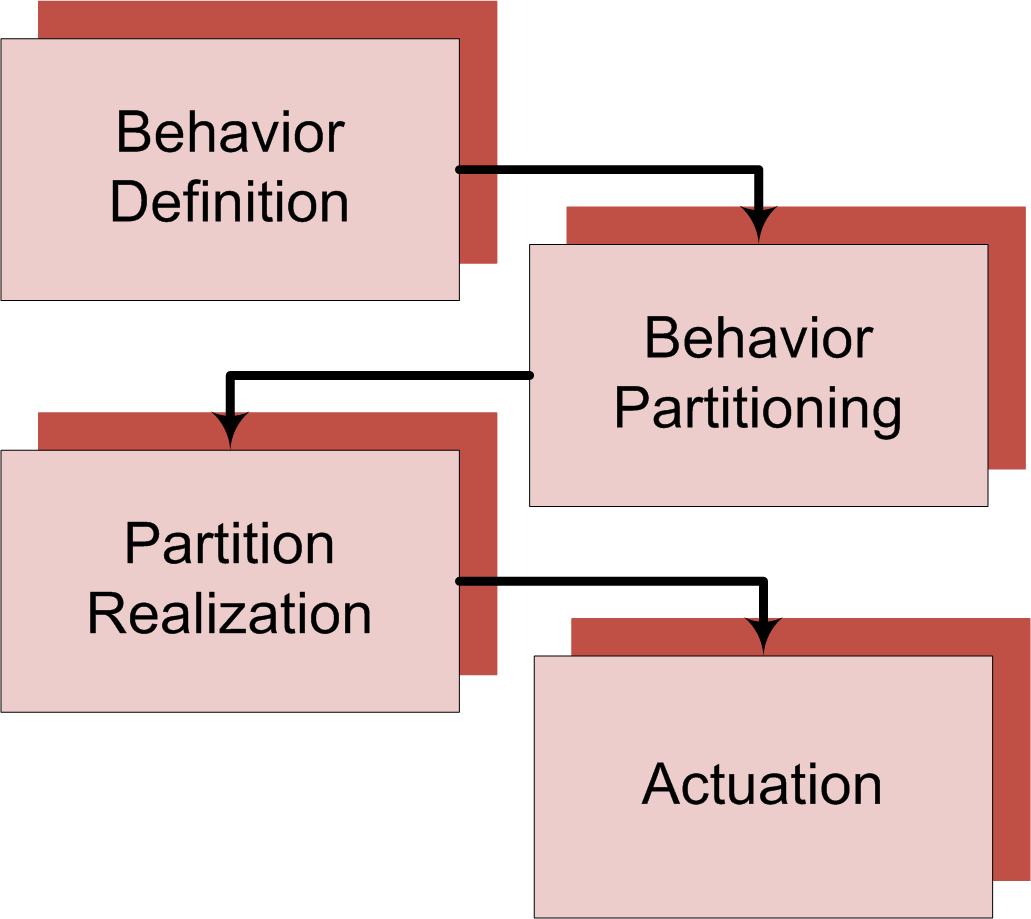}
\caption{Development Process from Desired Behavior to the Equivalent Vehicular Actuation}
\label{fig:process}
\end{figure}

\section{Formal Specification of the Development Process}
\label{section_2}

The development process for engineering a typical cognitive BV shall be clarified by some sequential phases. These phases shall be considered, sequentially. Actually, each phase is responsible for both processing the input stemmed from its previous phase and preparing the applicable input for the next phase. Specification of the aforementioned phases shall presented, as below:

\begin{itemize}
\item \uline{\textbf{\textit{Behavior Definition}}}: The desired behavior, with which the BV is supposed to be animated, shall be determined. Literally, the behavior would be defined, according to the reaction of the vehicle because of the applied stimulation stemming from the other objects' motion in local environment.
\item \uline{\textbf{\textit{Behavior Partitioning}}}: The defined behavior shall be partitioned in a hierarchical fashion. The behavior would be decomposed to the modular partitions, could which be designed, implemented and maintained, independently. Literally, different aspects of the behavior could be studied, separately. For example, when one has been called behind, she can turn back and move her eyes ,simultaneously. These actions are, literally, decomposed modules of the defined overall behavior within the mind. Such kind of partitioning lets us analyze the complex behaviors by investigation on their simpler components.
\item \uline{\textbf{\textit{Partition Realization}}}: A SNN-based circuit shall be designed for each partitioned module, so that each module's perception function does comprehend some aspects of the required behavior by its corresponding neural circuit. Actually, cascade interconnection of the designed circuits shall represent the overall partitioned complex behavior.
\item \uline{\textbf{\textit{Actuation}}}: Functional action circuits shall be devised to realize the behavior. The corresponding functions will control the agent's wheels to acquire the behavior. The neural control law drives the robot such that it will mimic the desired behavior in presence of the assumed environmental stimulation.
\end{itemize}

\begin{figure}[!t]
\centering
\includegraphics[scale=0.3]{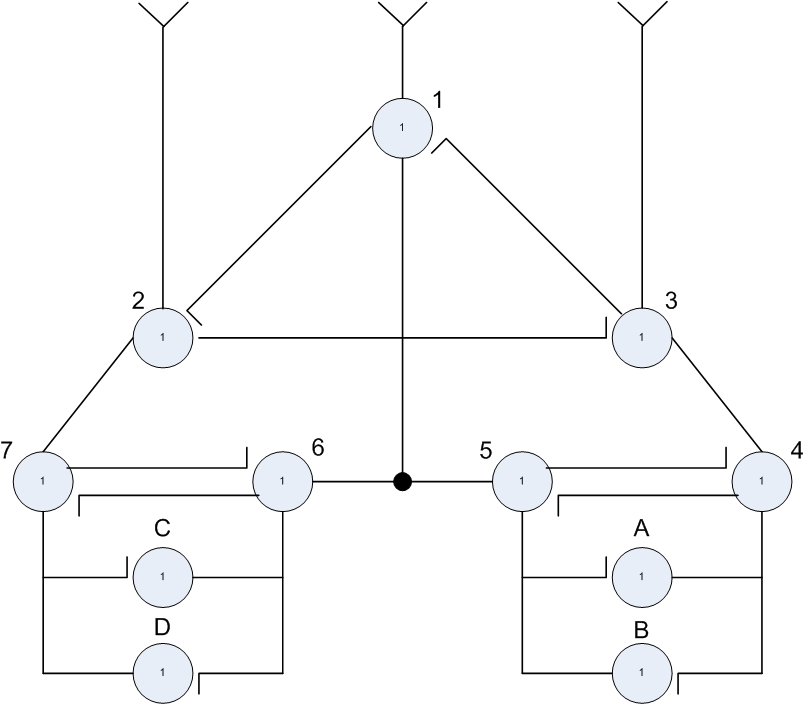}
\caption{Applied SNN into the Opportunist Vehicle to Realize the Planned Partitioned Modules [Excitatory and inhibitory connections have been shown by simple- \& crossed-line artifacts, respectively. The threshold of each neuron has been, also, noted as a number inside it.]}
\label{fig:circuit}
\end{figure}

This process, reasonably, could be known as a set of transformations from a complex behavior to simple behaviors, primarily. Then, each simple behavior would be mapped to its equivalent neural circuit. Finally, the aforesaid neural circuits would construct the necessary control law for the successful behavior replication. Sequential representation of the process would be depicted as Fig. \ref{fig:process}, graphically. Utilization of this process to design a BV, with due attention to specific behavioral requirements, is the subject of the next section.


\section{Opportunist: A Process-Oriented Formally-Developed BV}
\label{section_3}

\begin{figure}[!t]
\centering
\includegraphics[scale=0.15]{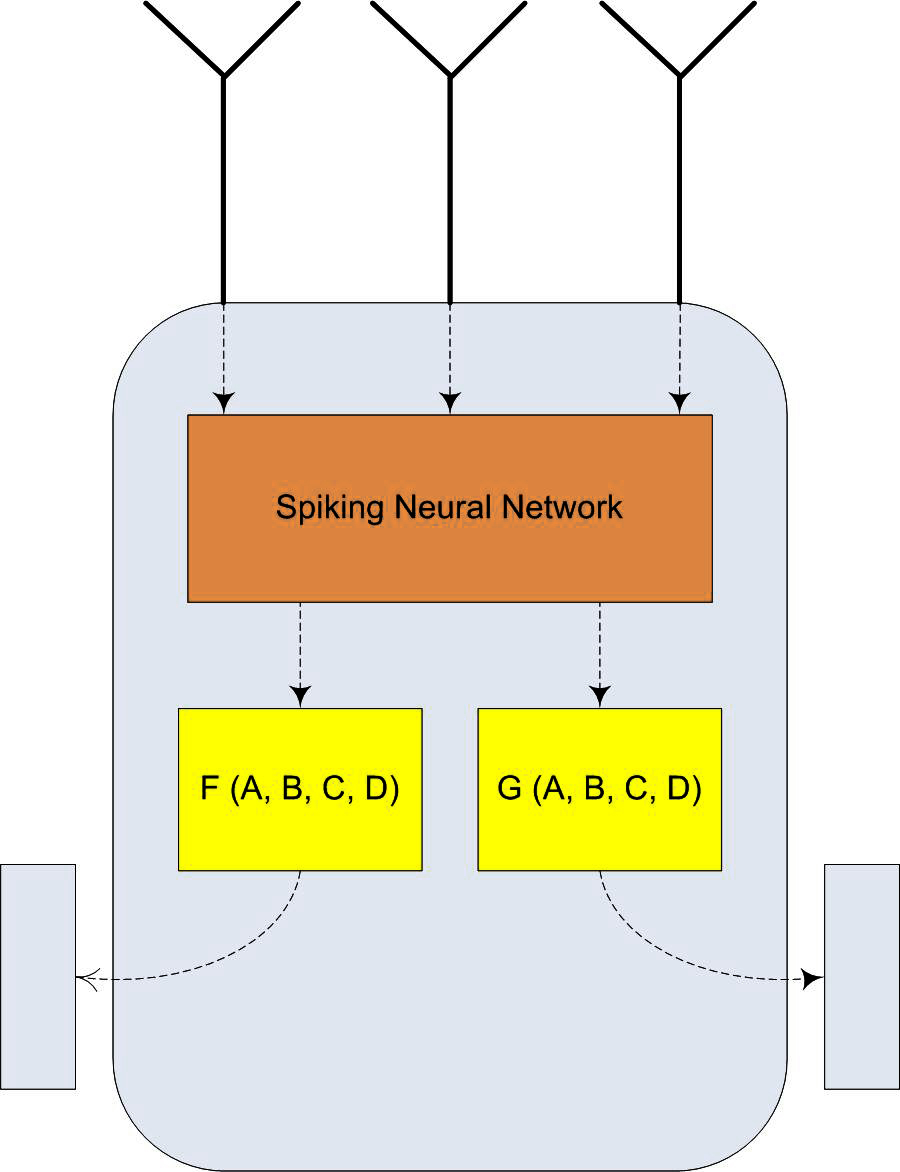}
\caption{Opportunist Vehicle's Explicit Realization}
\label{fig:vehicle}
\end{figure}

The design of a BV, according to above process, would be scrutinized, considerably, in this section. Based on the process, the desired behavior shall be determined, primarily. Typically, the cognitive capability to interact with the environment, suspiciously, shall be pursued. To this aim, the vehicle shall be able to follow a local agent. Furthermore, the vehicle will be conservative approaching to its mate and prohibit rapid approaching it. Moreover, any mate's aggressive motion toward the vehicle shall lead to vehicle's unconditional backtracking. The vehicle's endeavor is to exploit the following opportunity by chasing the mate, but not courageous enough, within interaction with it. Therefore, one could name this creature as "Opportunist", reasonably.

The vehicle should perceive the mate's proximity motions, qualitatively. The left-to-right(LR) or right-to-left(RL) motion would be independent of the approaching to distancing movements. So, the perception-driven aspect of the behavior shall be partitioned into these two coarse mechanisms. Actually, the vehicle shall detect the mate's horizontal motion direction, i.e. either LR or RL; Then, assessing the vertical quality of mate's motion.

Realization of the aforementioned modules could be realized by CTD mechanism \cite{6}, as shown in Fig. \ref{fig:circuit}. The upper module, consisting neurons $1$, $2$ \& $3$ will detects the horizontal direction, where the lower module, constructed from the rest of neurons, shall detect the depth of motion. Simultaneous utilization of excitatory and inhibitory connections among the agents has been accounted for minimization of the epileptic seizures within the operation of the neural circuits. Literally, simultaneous activation of a multitude of neurons will increase the overall potential of th circuit, leading to inappropriate power consumption, in a systematic perspective. Interested reader might investigate on \cite{5}, \cite{6} \& \cite{7} for more detailed elaborations about CTD mechanism, this circuit and the ideas behind its design.

Actuation planning shall be based on the CTD output, including neuron set $\{A,B,C,D\}$. Different desired motion scenarios, within which the vehicle must operate according to defined behavior, have been represented in the first column of Table \ref{table:logic}. Each row depicts the final potential of each neuron corresponding to its corresponding scenario. Literally, the last column shows the cumulative potential, could which be accounted for comparing quality of implementation for various scenarios, quantitatively.
The symbolic representation of Opportunist (Fig. \ref{fig:vehicle}), stipulates that the actuating control signals shall be generated by some logical functions, as $F(A,B,C,D)$ \& $G(A,B,C,D)$, and being applied onto the vehicle's wheels.

\begin{figure}[!t]
\centering
\includegraphics[scale=0.15]{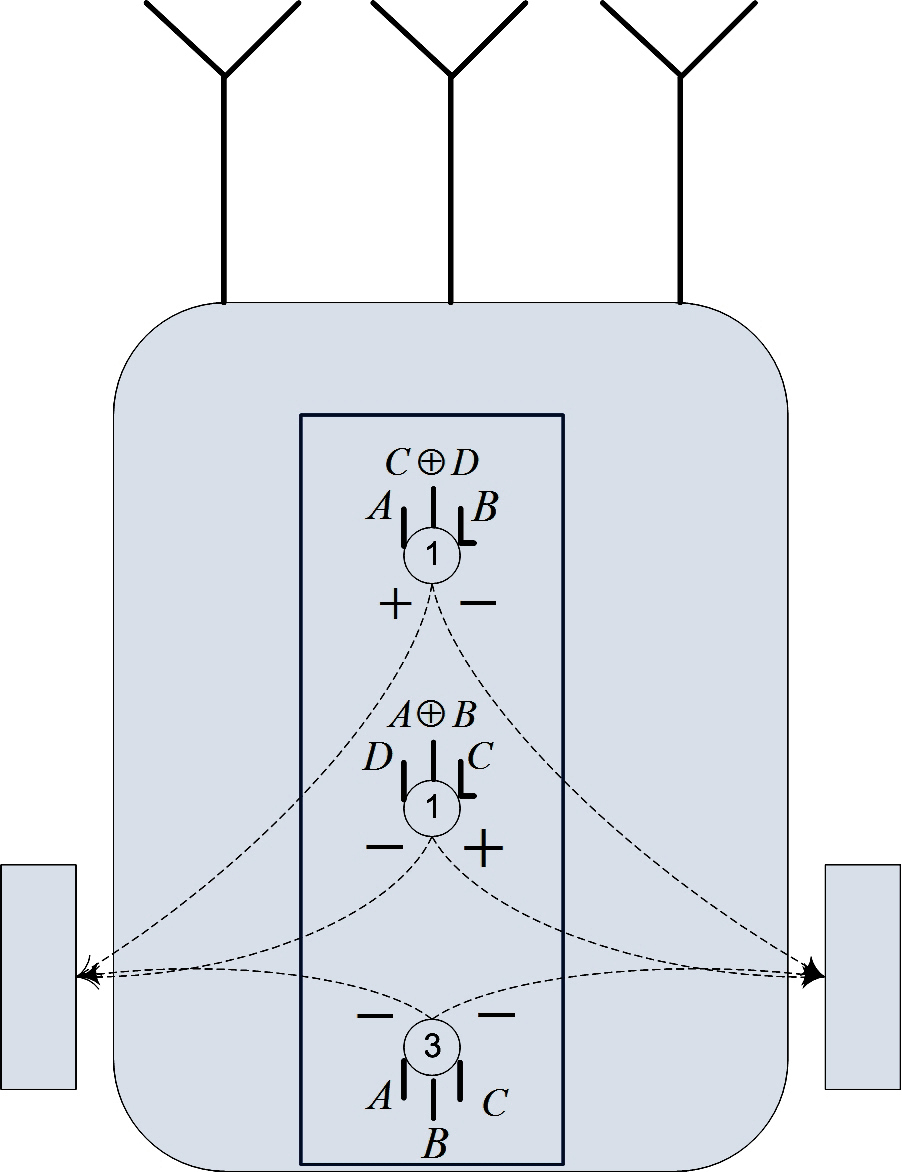}
\caption{Controlled Vehicle with the Derived Action Logic}
\label{fig:implementation}
\end{figure}

Table \ref{table:logic}'s control action logic could be applied onto the left and right wheels, logically, as specified below:

\begin{equation}
	\begin{cases}
	(L^{-},R^{-}): & ABC\\
	(L^{-},R^{+}): & \bar{C}D(A\oplus B)\\
	(L^{+},R^{-}): & A\bar{B}(C\oplus D)\\
	(L^{+},R^{+}): & O. W.\\
\end{cases}
\end{equation}

where each $(L^{\pm},R^{\pm})$ pair represents the state, within which the speed of left \& right wheels would be either increased or decreased, respectively, and $\oplus$ is exclusive OR operator. The implementation of the above control laws has been shown in Fig. \ref{fig:implementation}.

Following section shall unveil the behavior of this vehicle and assess the accomplishments, in comparison with the primary defined target behavior.


\section{Behavioral Simulations \& Analysis}
\label{section_4}

\begin{table}[!t]
\begin{threeparttable}

\renewcommand{\arraystretch}{1.3}

\caption{Neural Dynamics of the Vehicle's Desired Behavior}

\label{table:logic}
\centering
\begin{tabular}{c||c|c|c|c|c|c|c||c|c|c|c||c}
    & 1 & 2 & 3 & 4 & 5 & 6 & 7 & A & B & C & D & $\Sigma$\\
\hline
RLD &   & 1 &   & 1 &   &   & 1 & 1 &   &   & 1 & 5\\
\hline
RLS & 1 & 1 &   &   & 1 &   & 1 &   & 1 &   & 1 & 6\\
\hline
RLA & 2 & 1 &   &   & 2 & 1 & 1 & 1 & 2 & 1 & 1 & 6\\
\hline
\hline
LRD &   & 1 & 1 & 1 &   &   & 1 & 1 &   &   & 1 & 6\\
\hline
LRS &   &   & 1 & 1 &   & 1 &   & 1 &   & 1 &   & 5\\
\hline
LRA & 1 &   & 1 & 1 & 1 & 2 &   & 1 & 1 & 2 &   & 6\\
\end{tabular}

\begin{tablenotes}
      \small
      \item Applied nomenclature into the first column could be known as: R, L, D, S \& A stand for right, left, distancing, straight \& approaching, respectively. As an instance, RLD shall be interpreted as right-to-left distancing motion. The considered scenarios do cover all types of planar motion, would which stimulate the agent, exhaustively.
    \end{tablenotes}

\end{threeparttable}
\end{table}

The evaluation of the accomplishments, stemming from the novel development process, would be taken into account with simulation of the recently-derived BV, i.e. Oppurtunist. The described target behavior of the vehicle in section \ref{section_3} does, actually, establish a metric to assess our design and the process.

\begin{figure}[!t]
\centering
\includegraphics[scale=0.13]{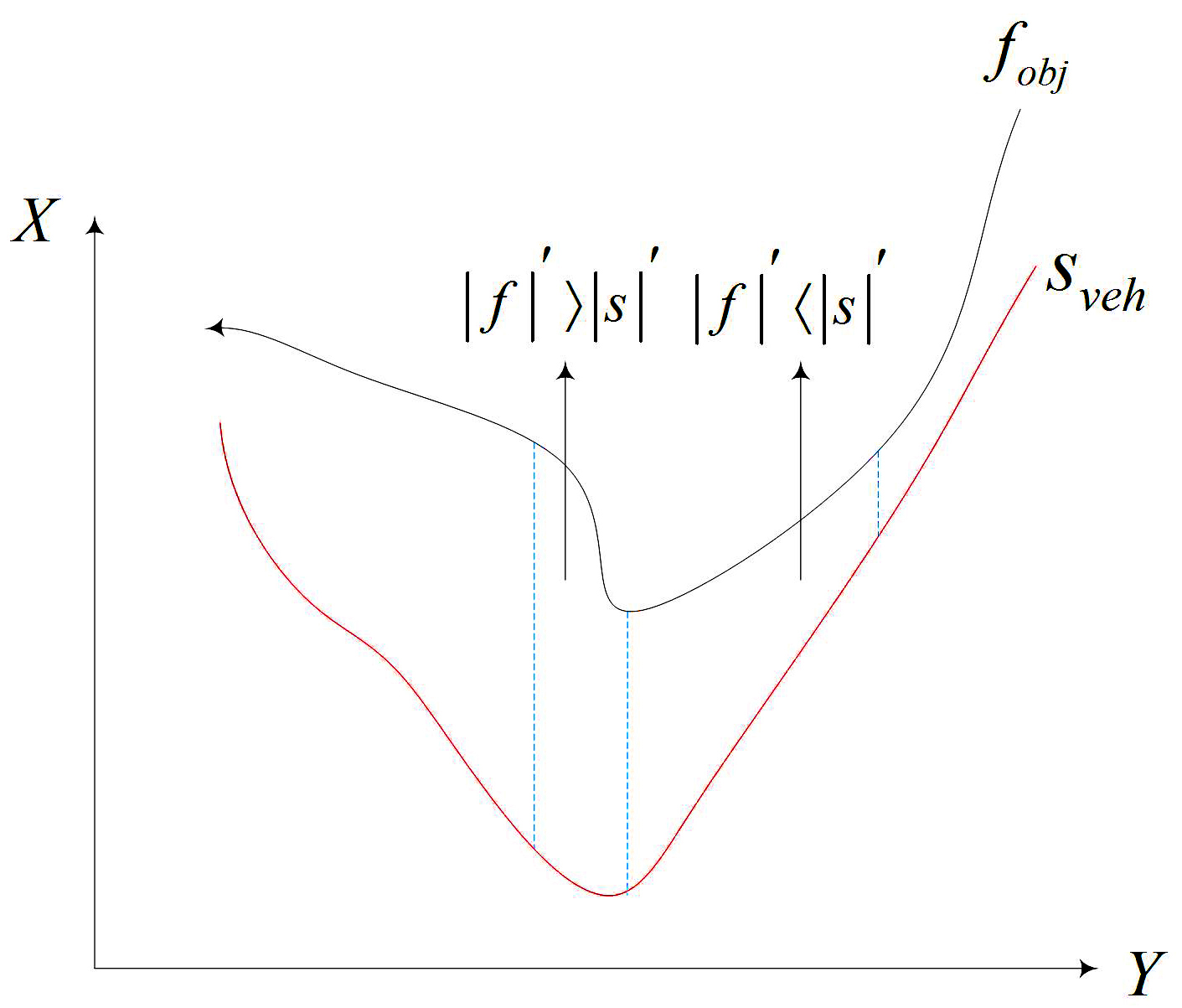}
\caption{Suspicious Dynamics of the Vehicle}
\label{fig:suspicious}
\end{figure}

The main characteristic behavioral property of Opportunist could be named as its desired policy to confront with the other agents. As described earlier, caution is expected to be embedded into the vehicle, as a behavioral attitude, when a typical mate does get away the vehicle. Furthermore, when a mate approaches the vehicle, its inherent hard-coded belief should consider that action as an aggressive one. So, it shall be supposed to stray from its trajectory, at least temporarily. Simulated dynamics of this suspicion-based behavior has been depicted in Fig. \ref{fig:suspicious}. Note that within all of the forthcoming simulations, target vehicle's trajectory would be shown by red color, is which being traversed due to its mate's motion. Furthermore, one could discern the mate's trajectory in black color.

Assume $f_{obj}$ and $s_{veh}$ as the trajectory functions of the mate and the vehicle, respectively. Whenever the mate approaches the vehicle, the vehicle starts to get away, fast, as it has been designed to consider such stimulation as an aggressive perception. Therefore, its escape has been, deservedly, entails the desired behavior, as  $\lvert f_{obj}\rvert^{'} < \lvert s_{veh}\rvert^{'}$. As the mate, conversely, gets away the vehicle, the vehicle should approach the mate according to the behavioral requirement, albeit with the expected suspicion. So, it is inevitable that the vehicle's approaching rate would be lower than its formerly-explained escaping tendency. This rest of the shown trajectory in Fig. \ref{fig:suspicious} proves the credible implementation of this behavior, as $\lvert f_{obj}\rvert^{'} > \lvert s_{veh}\rvert^{'}$

\begin{figure}[!t]
\centering
\includegraphics[scale=0.13]{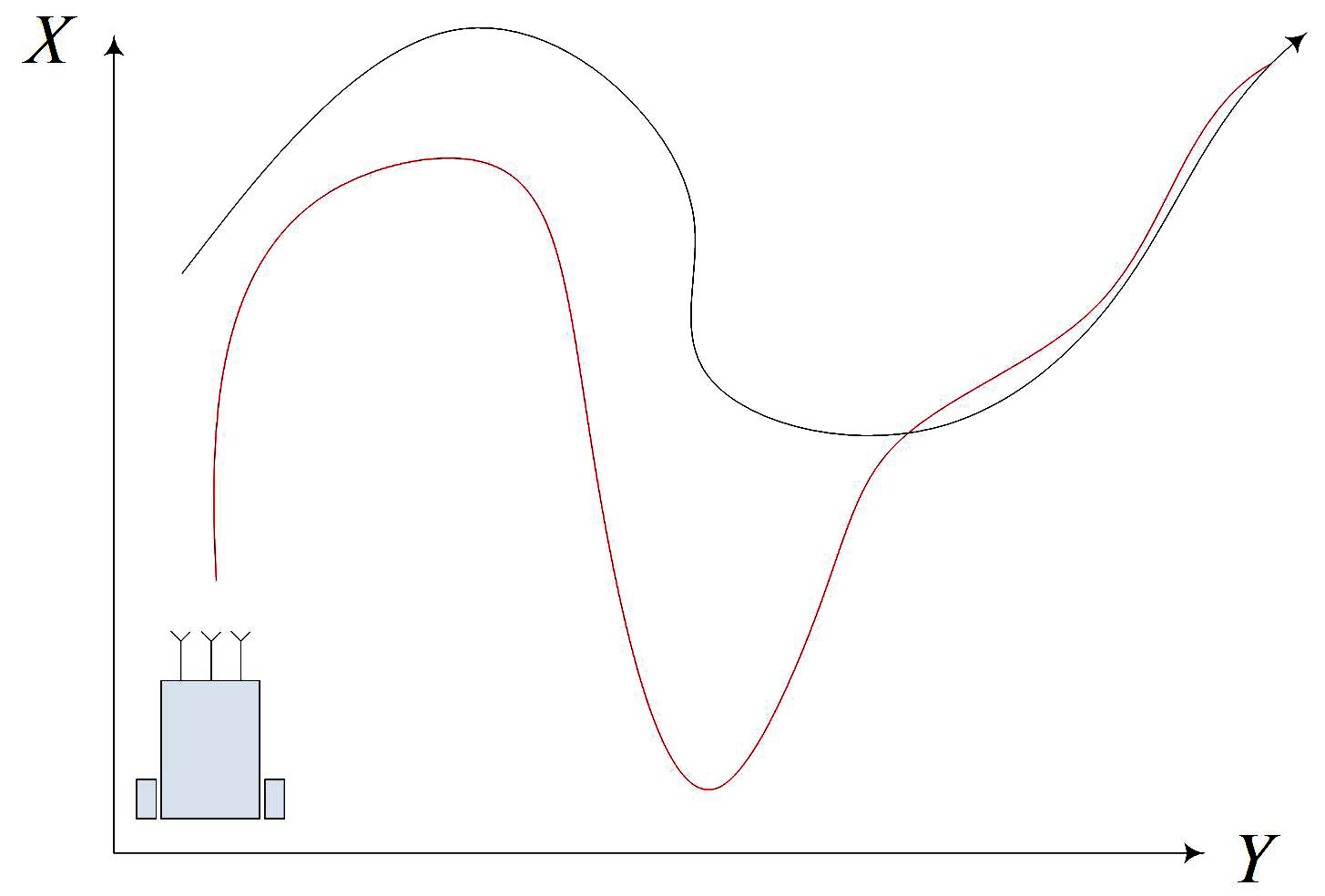}
\caption{Vehicle's Following Behavior}
\label{fig:following}
\end{figure}

The other aspect of the required behavior would be the efficiency of the vehicle's following capability, should which be syncretized with suspicion. A specific simulated scenario has been run with a longer trajectory, comparing with the former. According to Fig. \ref{fig:following}, the vehicle approaches the mate and tries to be opportunist and follow it, as long as it has shown no aggressive behavior. Upon the emergence of early witnesses for approaching the mate, the vehicle will cease to follow the mate and stop chasing it. As soon as, the mate changes its direction, the vehicle turns back and chase it as a typical successful following maneuver. On could, truly, claim on the noticeable quality of the trajectory, in view of the steepness and suspicion-driven behavior of the vehicle.

Third scenario challenges the Opportunist in perspective of a long-running wandering simulation, is which in presence of a mate. The mate, as usual, would be considered as the stimulation source for the vehicle. In this test, the mate's trajectory has been planned to be a relatively-circular path to let the observer discover the vehicle's behavioral aspects, have not been unveiled by former simulations, yet. With due attention to acquired trajectories, the vehicle's loyalty to suspicion and opportunistic policy have been proved. In different phases of the simulation, the vehicle's endeavor has concentrated on following the mate. whenever the mate changes its direction toward the vehicle, it would withdraw from chasing the mate and deviation in trajectory shall be happened. As the mate continues either its former type of traversing or increasing its distance from the vehicle, it would try to follow the mate, subsequently.

The Opportunist's behavior could, not only, be verified by the analysis of its interactions, similar with the above approach, but also with the possible backtracking from the product (the action circuit) to the primary requirements (the defined target behavior). Applied bijective mapping, has which been planned for each phase of the process with its former and latter phases, lets manipulate the design to find new types of the expected potential behaviors for a vehicle.


\section{Concluding Remarks \& Future Works}
\label{section_5}
This research's contribution is a common ground for engineering neural reactive agents in a fully-process-driven manner. This strategy fulfills the acquisition of adaptivity, does which empower the vehicle to react to diverse environmental stimulations, properly. It is worth noting that the typical BVs just animates one behavioral pattern, solely. The applied partitioning phase would handle the complicated design tasks by segregation of the different levels of behavior, hierarchically. Partition realization phase does, noticeably, conduct the design efforts to optimal mapping of the desired behavior to the responsible BV, under the aegis of behavior partitioning. One might comprehend the importance of the verification process within this paper, whose possibility has been taken into account within this manner of formal development. Actually, the demonstrated approach facilitates the verification process of a typical agent and the bidirectional traceability between the primary behavior and the agent's behavioral outcomes. Finally, engineering a cognitive BV by this methodology and evaluation of the denouements stressed on the claimed credibility of the process.

This development framework would initiate some trends of research to strengthen its application scope. The framework is able to realize the behavior for one agent, where the approach would be extendable to a team of robots. Actually, local behavior assignment to global multi-agent systems might ignite a research stream to generalize the formal development process. Opportunist vehicle may have other behaviors, where the perceived object's speed will vary, considerably. An analytic work would be necessary to describe a comprehensive overview from its behavior. Hence more cognitive functionalities would be expected to be observed from this complicated BV.

Epileptic analysis of SNNs should, also, be addressed in neural circuits. With increasing complexity of the applied behaviors, the circuits will grow, such that typical rough assessment of potential would not be possible, efficiently. This process could be mobilized with inline procedures to evaluate the possibility of epileptic seizures within the planned circuits, leading to optimization of the cross-correlation among connections and realization of efficient circuits with reliable performance in a wide range of spiking frequencies.

\addtolength{\textheight}{-15cm}   

\begin{figure}[!t]
\centering
\includegraphics[scale=0.13]{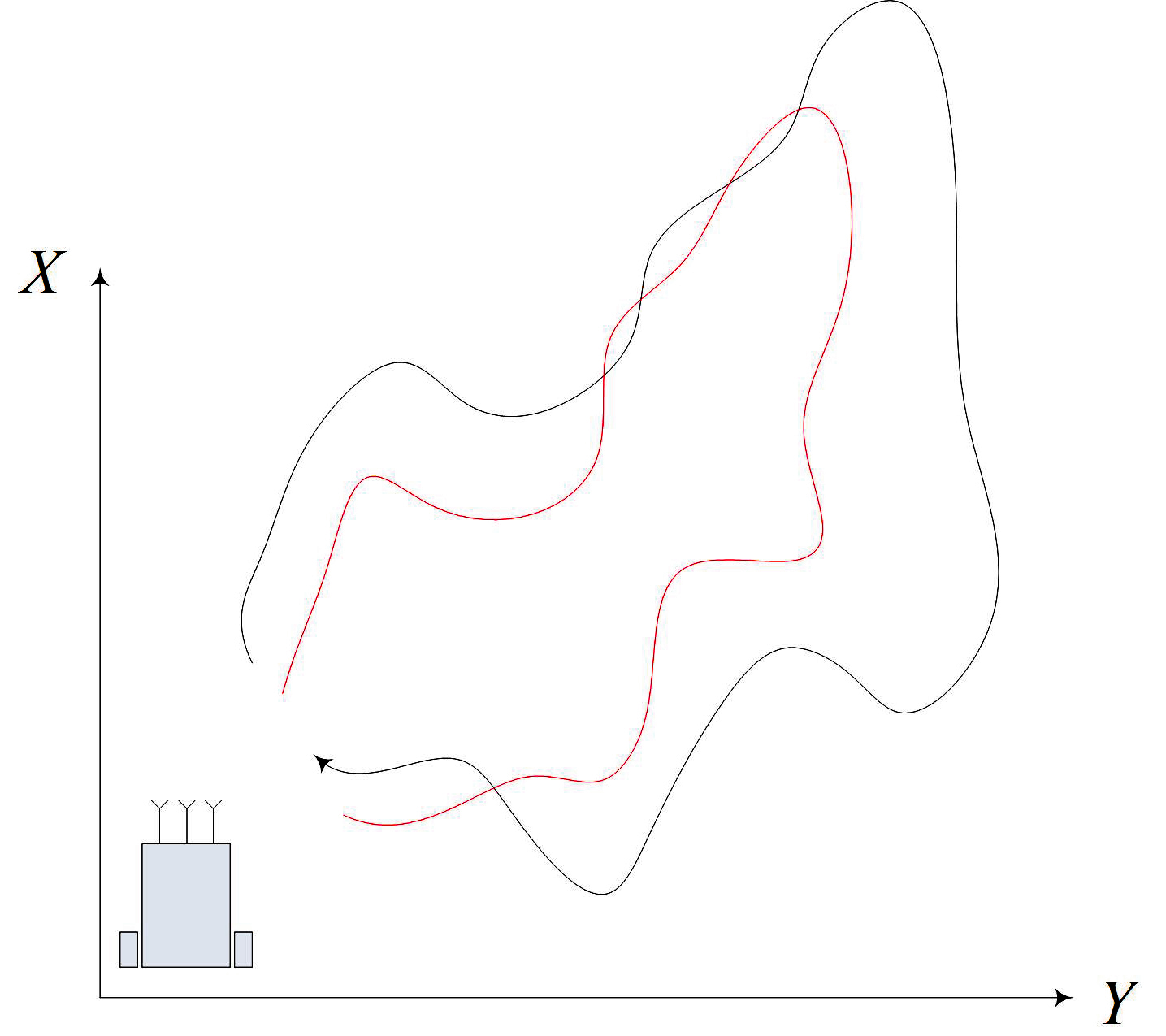}
\caption{Vehicle's Wandering Behavior}
\label{fig:wandering}
\end{figure}





\section*{Acknowledgment}
Special thanks Nasrin Shirali, Elnaz Ghezelbash, Yasin Gorgij  \& Erfan Nasoori for their endeavors to distribute process-oriented knowledge, thinking \& engineering, as the central inspiring attitude, does which govern this research.

\end{document}